\documentclass[10pt,twocolumn,romanappendices]{IEEEtran}

\usepackage[noadjust]{cite}
\usepackage[numbers,square,comma,sort&compress]{natbib}
\usepackage{calc}
\usepackage{comment}
\usepackage{graphicx}
\usepackage{amsmath,epsfig,amssymb,verbatim,amsopn,cite,subfigure,multirow}
\usepackage{balance}
\usepackage[usenames,dvipsnames]{color}
\usepackage[all]{xy}  
\usepackage{url}
\usepackage{amsfonts}
\usepackage{amssymb}
\usepackage{epsfig}
\usepackage{slashbox}
\usepackage{bm}
\usepackage{epsfig}
\usepackage{amsmath,amssymb, amstext}
\usepackage{graphicx}
\usepackage{epstopdf}
\usepackage{algorithm}
\usepackage{algorithmic}
\usepackage[numbers,square,comma,sort&compress]{natbib}
\allowdisplaybreaks

\begin{document}

\title{\huge A Novel Low-Complexity Framework in Ultra-Wideband Imaging for Breast Cancer Detection}


\author{Yasaman~Ettefagh$^\dag   $, Mohammad Hossein~Moghaddam,$^* $ and Saeed~Vahidian$^\#$\\ 
\small
$^\dag$ Department of Electrical Engineering, Amirkabir University of Technology, Tehran, Iran\\
Email: \{yasaman.ettefagh@aut.ac.ir\} \\

$^* $ Department of Electrical Engineering, K.N. Toosi University of Technology, Tehran, Iran\\
Email: \{mhmoghadam@ee.kntu.ac.ir\} \\

$^\#$ Department of Electrical and computer Engineering, University of Illinois at Chicago, Chicago, USA\\
Email: \{svahid2@uic.edu\} \\

}
\normalsize
%


\maketitle

\begin{abstract}
In this research work, a novel framework is proposed as an efficient successor to traditional imaging methods for breast cancer detection in order to decrease the computational complexity. In this framework, the breast is devided into segments in an iterative process and in each iteration, the one having the most probability of containing tumor with lowest possible resolution is selected by using suitable decision metrics. After finding the smallest tumor-containing segment, the resolution is increased in the detected tumor-containing segment, leaving the other parts of the breast image with low resolution. Our framework is applied on the most common used beamforming techniques, such as delay and sum (DAS) and delay multiply and sum (DMAS) and according to simulation results, our framework can decrease the computational complexity significantly for both DAS and DMAS without imposing any degradation on accuracy of basic algorithms. The amount of complexity reduction can be determined manually or automatically based on two proposed methods that are described in this framework.
\end{abstract}


\section{Introduction}

\IEEEPARstart{B}{reast} cancer represents the second most common type of cancer in women after lung cancer \cite{colgan20153}. The current detection method is mammography which uses X-rays to detect the cancerous tumor. Mammography has several drawbacks which has encouraged researchers to look for other alternatives. Using X-rays, makes the test ionizing, so mammography itself can increase the risk of breast cancer. So while the early detection is very crucial for full recovery, mammography is not recommended more than once or twice a year. Moreover, there is a small difference between the electrical properties of malignant and benign tissues in X-rays frequencies, resulting in 1. high false-negative rate (4-34\%) and 2. high false-positive rate (70\%)\cite{huynh1998false ,elmore1998ten}. Also, mammography is involved with breast compression which makes it unpleasant for patients. Microwave imaging is a suitable alternative for mammography which 1. is not ionizing so several tests can be done annually which would increase the chance of early detection without causing any risk and 2. is more discriminating since the difference between electrical properties of malignant and benign tissues in microwave frequencies is much more intense than X-rays frequencies.

In recent years, several microwave imaging algorithms have been proposed for breast cancer detection \cite{bond2003microwave,lim2008confocal,o2010quasi}.
In this field, one of the most successful set of methods is confocal microwave imaging methods. Two of the basic methods in this set of algorithms are delay and sum (DAS) \cite{bond2003microwave} and delay multiply and sum (DMAS) \cite{lim2008confocal}. These methods are very simple for implementation, but are efficient only for simple test data. In case of considering clutter and skin effects, these algorithms fail to incept the tumor. More intelligent confocal imaging-based methods which can detect tumor in presence of clutter and skin effects have been proposed \cite{jalilvand2011uwb,byrne2015time}.
Generally, radar imaging methods are classified into two sets: data independent (DI) algorithms and data adaptive (DA) algorithms. The weights of DI algorithms are fixed regardless of the dataset, while weights of DA algorithms are adjusted according to the second order statistics of dataset. DAS and DMAS, are among simple DI algorithms and in case of more efficient, but more complicated algorithms, MIST has been proposed \cite{bond2003microwave}.
Data adaptive nature of DA algorithms, enables them to mitigate the effect of noise and clutter in a more efficient and adaptive way. As an example, the Capon beamformer, also termed the Minimum Variance Distortionless Response (MVDR) method, calculates beamformer weights to minimize the output power across the array while limiting the beamformer response to allow signals from the desired direction to pass with specified phase and gain \cite{harry2002optimum}.
Based on the particular method of collecting backscatter signal, the imaging setup can be monostatic, bistatic or multistatic. In monostatic setup, the transmitter and receiver are realized in a single antenna. In bistatic setup, transmitter and receiver are separated and multistatic setup is composed of multiple antennas each being both transmitter and receiver.

In this research work, a novel framework for computational complexity reduction of beamforming techniques have been proposed which can be applied to DI algorithms such as DAS and DMAS methods. In our setup, the multistatic imaging with a ring of 12 Vivaldi antennas is used for two different datasets with different tumor locations and different breast phantom sizes and both DAS and DMAs algorithms are used in to the test for our framework. According to simulation results, our method can decrease the computational complexity of DAS in a scale of up to 1:5 and complexity of DMAS of up to 1:13 without any loose of accuracy on tumor detection.

The rest of this paper is organized as follows: In section II we go through a brief description of DAS and DMAS approaches. In section III the proposed low complexity framework is described. Section IV presents the simulation results, and section V concludes this paper.

\section{Literature Review}
In this section, we have a brief review on the two most basic beamforming techniques: DAS and DMAS. Our proposed method will be implemented as a framework on these two techniques.

\subsection{DAS Method}

Delay and sum is the most basic beamforming technique. In this technique, a number of antennas are located around the breast each acting as transmitter or receiver depending on the specific data collecting method which could be monostatic or multistatic. The breast tissue is divided into several focal points according to the desired resolution. For each focal point and backscatter signal from a specific pair of transmitter and receiver, the roundtrip path lengths among the transmitter, focal point and the receiver is calculated and converted into time delay. This delay is applied to the received backscatter signal and this process is repeated for all backscatter signals from each pair of transmitters and receivers. The amount of delay for monostatic system is given in (1) and for multistatic systems in (2).

\begin{equation}
{n_i}(\overrightarrow r ) = \frac{{|\overrightarrow r  - \overrightarrow {{r_{iT}}} |}}{{2 * c * \Delta t}}
  \end{equation}

\begin{equation}
{n_{ij}}(\overrightarrow r ) = \frac{{|\overrightarrow r  - \overrightarrow {{r_{iT}}} |}}{{c * \Delta t}} + \frac{{|\overrightarrow r  - \overrightarrow {{r_{jR}}} |}}{{c * \Delta t}}
  \end{equation}

where $\overrightarrow r$, $\overrightarrow {{r_{iT}}}$  , $\overrightarrow {{r_{jR}}}$ , c and $\Delta t $ denote the location of focal point, location of the ${i^{th}}$ transmitter, location of the ${j^{th}}$ receiver, the speed of light and sampling time respectively.

Finally, we sum up all aligned backscatter signals for each focal point and we repeat these steps for all focal points. So a plot containing synthesized energy of all focal points is obtained using (3) for monostatic systems and (4) for multistatic systems. Since tumor is a scatterer object, we expect the synthesized energy for tumor-containing focal points to have maximum value which indeed is the case. So in this way location of the tumor will be detected.

\begin{equation}
I(\overrightarrow r ) = \sum\nolimits_{t = {t_0}}^{{t_{N - 1}}} {\left[ {{{(\sum\nolimits_{i = 1}^M {{Y_i}(t - {n_i}(\overrightarrow r ))} )}^2}} \right]}
  \end{equation}

\begin{equation}
I(\overrightarrow r ) = \sum\nolimits_{t = {t_0}}^{{t_{N - 1}}} {\left[ {{{(\sum\nolimits_{i = 1}^M {\sum\nolimits_{j = 1}^M {{Y_{ij}}(t - {n_{ij}}(\overrightarrow r ))} } )}^2}} \right]}
  \end{equation}

where M is the number of antennas, $Y_i$ is the received signal in monostatic setup and $Y_{ij}$ is the received signal in multistatic setup.

\subsection{DMAS Method}
In Delay Multiple and Sum (DMAS) algorithm \cite{lim2008confocal}, in order to have much robust output, each pair of aligned signals are multiplied and then are summed up. DMAS has better image quality with the expense of higher computational complexity. The algorithm is shown in Fig.1 \cite{lim2008confocal}.

\begin{figure}[h]
  \noindent
 \raggedleft
  \noindent
  \includegraphics[width=8cm]{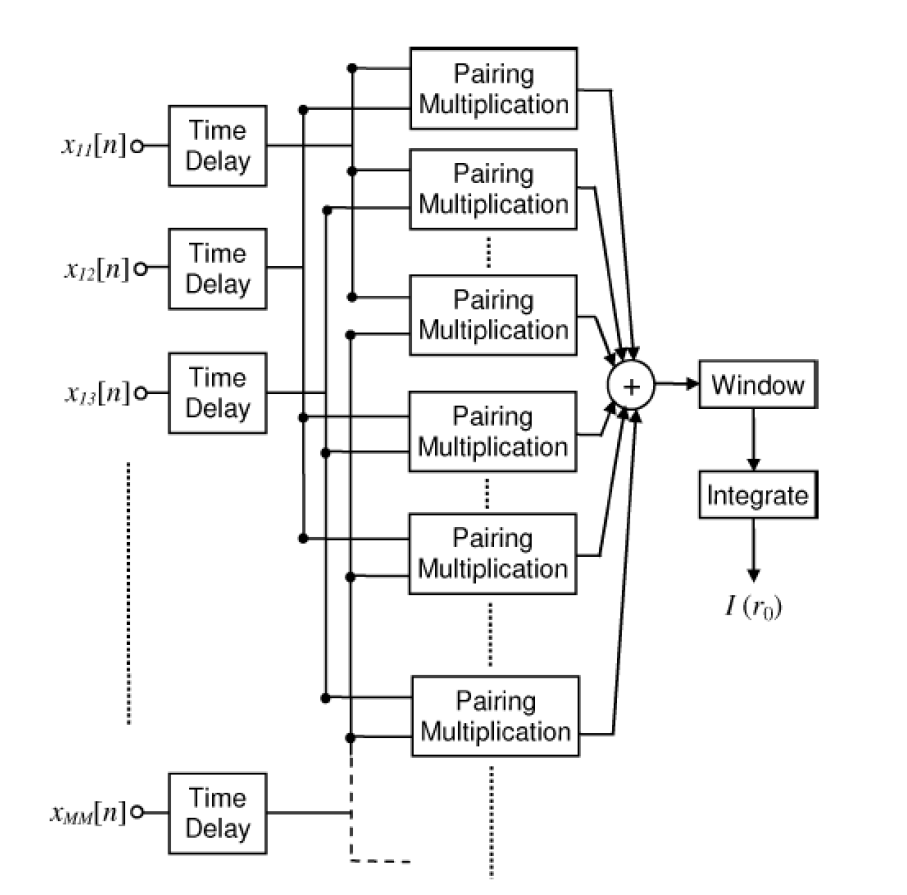}{\centering}
  \caption{DMAS Algorithm \cite{lim2008confocal}}
\end{figure}

\section{Proposed Computational Complexity Reduction Approach}

In beamforming algorithms, the breast tissue is divided into several focal points according to the desired resolution and apply the algorithm for all of them for which most of them will not indicate tumor so will be out of interest. Actually, there will be lots of unnecessary high spatial resolution calculations since we are only interested in finding the location of the tumor. In this paper, we propose a framework by which these unnecessary calculations are avoided. The idea is that the smallest possible region of interest (ROI) which has the most probability of having tumor will be identified by using few calculations with low number of focal points. Then, the number of focal points will be increased to the desired resolution and the algorithm make calculations for all of the increased focal points just in most suspicious ROI. For this purpose, a spatial decimation factor is defined, determining the number of reduced focal points for which the initial phase of beamforming algorithm is calculated. More spatial decimation values result in more reduction in computational complexity and on the other hand, also more more risk of missing the tumor-containing region.


In our proposed framework, the beamforming technique is applied on just a decimated number of focal points within the breast. We divide the breast into 4 segments and use a decision metric for selecting the segment with the most probability of having tumor. This segmentation and selection on the designated segments continued until 2 distance metrics are satisfied. We divide our framework into 3 phases:

\begin{itemize}
  \item Spatial Decimation

  First, we select limited number of focal points and apply beamforming algorithm on them. We use a parameter called spatial decimation factor and apply a linear sampling on original set of focal points with respect to the final resolution. Spatial decimation factor can be selected in two different modes:

\begin{itemize}

  \item Automatic mode in which spatial decimation factor is selected with respect to the smallest size of desired detectable tumor as an input. The efficiency and precision of our algorithm can be adjusted by setting this parameter. If we are about to detect bigger tumors, we can set this parameter to a pretty large value so more reduction in computational complexity will be achieved and vice versa.
  \item Manual mode in which spatial decimation factor is selected manually. No constraint is applied so we can have desired reduction in computational complexity and on the other hand, there would be risk of missing tumor location as well.

\end{itemize}

  \item ROI Selection

  Now we have the synthesized energy profile of the decimated number of focal points within the breast. We divide the breast into 4 equal squares and calculate variance of the synthesized energy in each squares and the whole region as well. We define a decision metric and nominate the tumor containing square and repeat this step until two distance metrics are satisfied. This decision metric could be mean or variance and we defined an intuition based on a decision metric as following:
  \begin{equation}
{M_i} = (1 - (\frac{{\sigma _i^2 - \sigma _{i - 1}^2}}{{\sigma _i^2}}))*{\mu _i} + (\frac{{\sigma _i^2 - \sigma _{i - 1}^2}}{{\sigma _i^2}})*\sigma _i^2
  \end{equation}
  in which, $\sigma _i^2$ is the variance of intensity points of selected segment in step $i$ and ${\mu _i}$ represents the mean of intensity points of selected segment in step $i$. In this way, when the variance in current and previous steps,  the dominant term would be mean (${\mu _i}$) and when the variance in current and previous steps have a
distinct difference, the dominant term would be variance $\sigma _i^2$.
  In the first iteration, we expect the tumor to be located in one of these squares together with some other tumor-free focal points. So regarding the high energy dispersion in the tumor containing square, we expect this square to have higher variance than other squares. So this square is selected as the tumor-containing one. As we approach the tumor, we expect to have more concentrated energy in tumor-containing square, so it seams that mean is becoming a better decision metric. We use within class distance and between class distance for termination condition as below:

  \begin{equation}
W = \sum\nolimits_{i = 1}^K {\sum\nolimits_{j = 1}^{{N_i}} {\left( {{x_{ij}} - \overline {{x_i}} } \right)} } {\left( {{x_{ij}} - \overline {{x_i}} } \right)^T}
  \end{equation}

\begin{equation}
B = \sum\nolimits_{i = 1}^K {{N_i}} \left( {{{\overline x }_i} - \overline x } \right){\left( {{{\overline x }_i} - \overline x } \right)^T}
  \end{equation}

  Here by classes, we mean two consecutive iterations. While we have increase in between class distance and decrease in within class distance, this step will continue.

  In order to remove the selection error for tumors which are selected on the edges of our divided squares in each iteration, we apply a sliding framing on the original boundaries of the selected ROI proportional to its length to expand it so it would contain the probable existing tumor on the edges.

  \item Final Imaging

  The smallest possible ROI has been selected. Now, we increase the number of focal points according to the desired resolution in the designated square. So the final image will have one region with high spatial resolution and the rest with low spatial resolution.
\end{itemize}


\section{Numerical Experiments}
\vspace{-1mm}

We test our framework on two sets of data for two different breast phantom sizes and two different tumor positions, considering a multistatic imaging setup consisting of a ring of 12 Vivaldi antennas. First dataset has a 1cm tumor at [-2,-3,0] with ${\varepsilon _r} = 50$ in a breast phantom with radius of 10cm, and second dataset has a 1cm tumor at [2,0,1] with ${\varepsilon _r} = 50$ in a breast phantom with radius of 5cm. DAS and DMAS, are used as the reference beamforming techniques considering no clutter and skin effects. In order to detect a supposed circular tumor with a particular size, we determine the spatial decimation factor with respect to the dimension of the inserted square in it. For a 1cm circular tumor, the dimension of the inserted square would be 0.7 cm. So for each 0.7cm, we need to select at least one point in order to not miss a 1cm tumor. So regarding the desired resolution which is 1mm, the spatial decimation factor would be 7.

In figure 2, we show the simulation results of basic DAS/DMAS and automatic/manual DAS/DMAS for two datasets. The amount of computational reduction is much apparat in DMAS which has more calculations in the original algorithm as well. In terms of elapsed time, we can see a 13 times reduction in complexity for DMAS algotirhm.

\begin{figure}[h]
  \noindent
 \raggedleft
  \noindent
 \includegraphics[width=9cm]{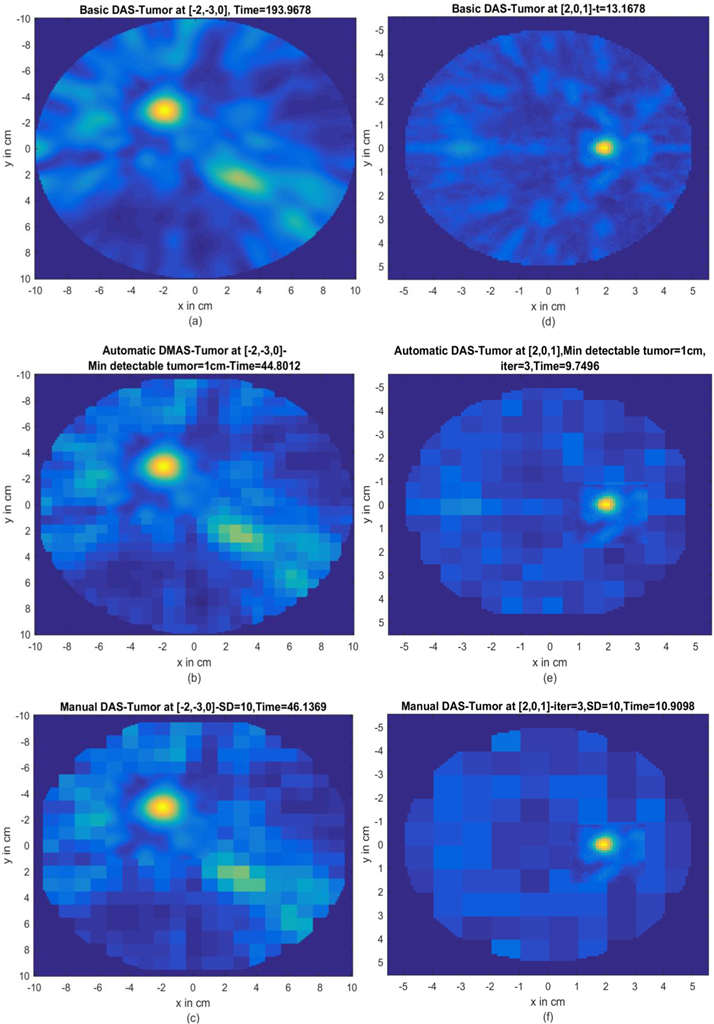}{\centering}
  \caption{Performance comparison of basic DAS ((a) and (d)) , manual DAS ((b) and (e)) and automatic DAS ((c) and (f)) for dataset1 (tumor at [-2,-3,0] (a)-(c)) and dataset2 (tumor at [2,0,1] (d)-(f))}
\end{figure}

\begin{figure}[h]
  \noindent
 \raggedleft
  \noindent
 \includegraphics[width=9cm]{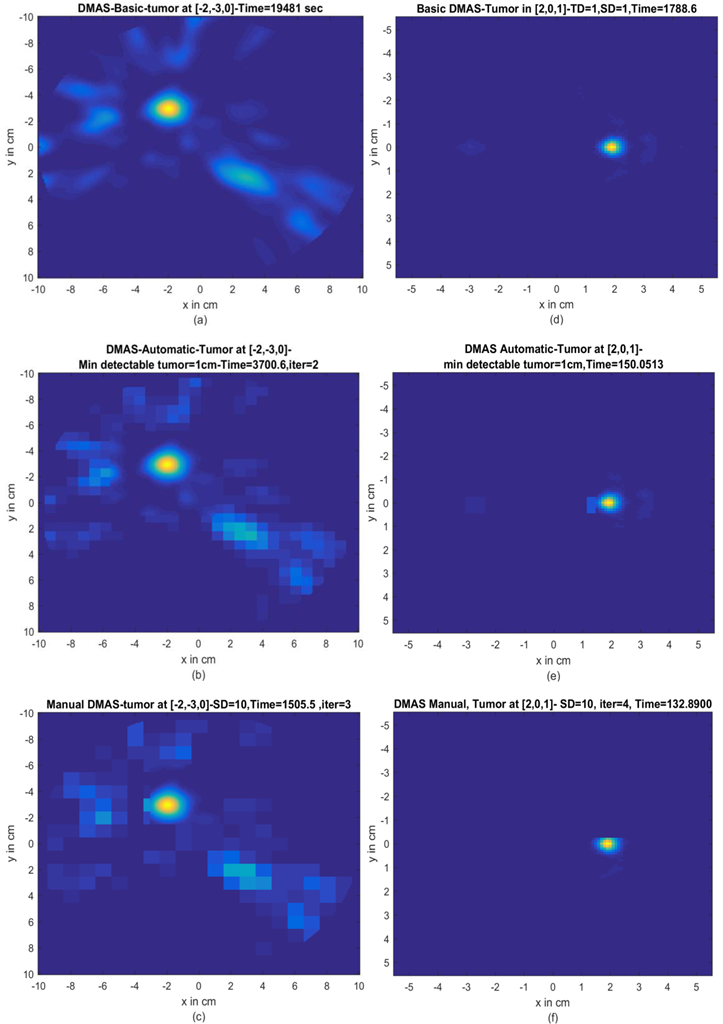}{\centering}
  \caption{Performance comparison of basic DMAS ((a) and (d)) , manual DMAS ((b) and (e)) and automatic DMAS ((c) and (f)) for dataset1 (tumor at [-2,-3,0] (a)-(c)) and dataset2 (tumor at [2,0,1] (d)-(f))}
\end{figure}

The performance of these simulations which include elapsed time and number of iterations is summarized in table I. Number of iterations, is the number of segmentations in ROI selection phase.

\begin{table}[H]
\begin{center}
\caption{Performance comparison of basic DAS/DMAS, manual DAS/DMAS and automatic DAS/DMAS in terms of elapsed time and number of iterations}
\includegraphics[width=9cm]{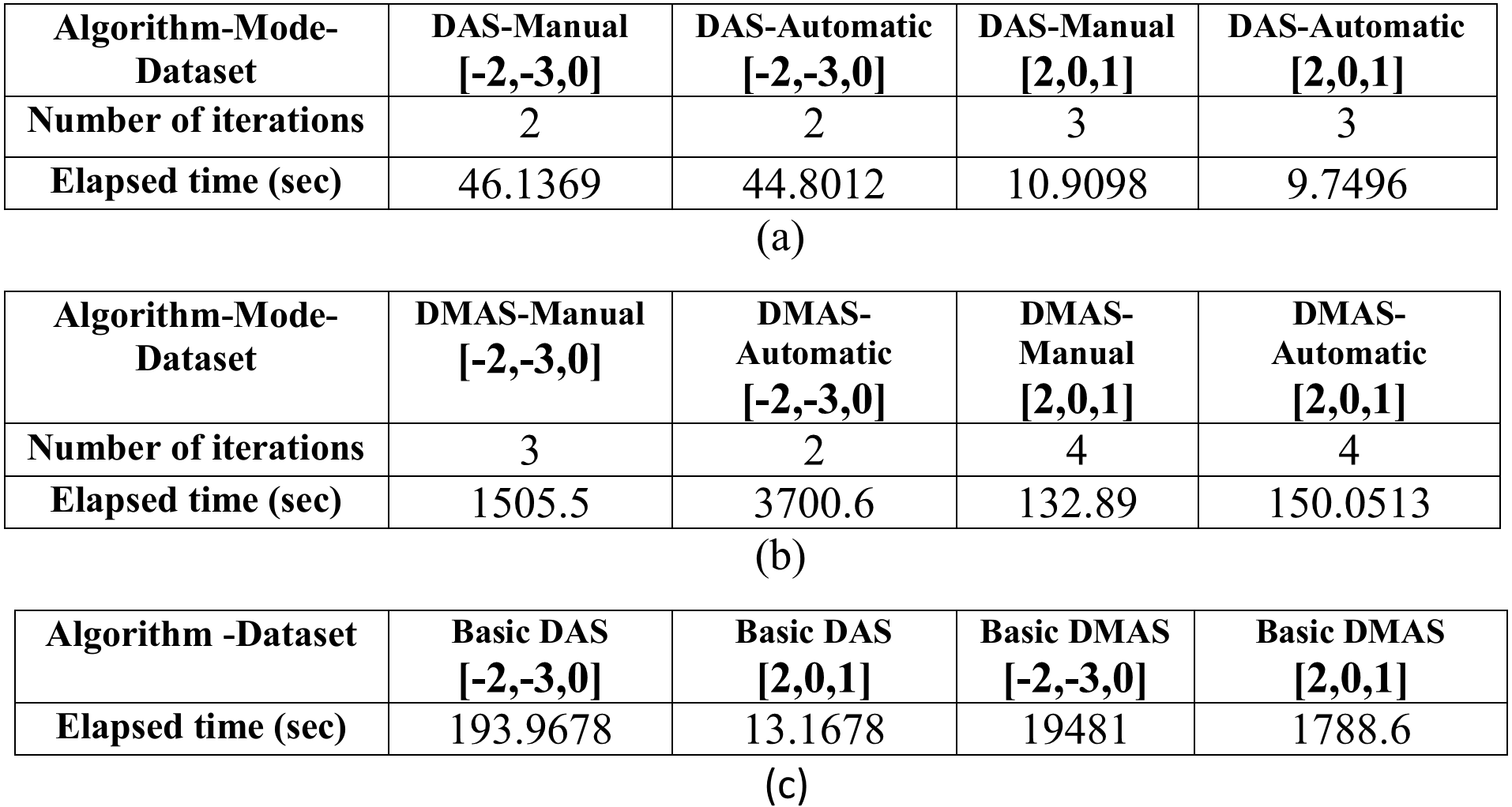}
\end{center}
\end{table}
In our study we supposed that tumor exists and we are about to locate it, but for non tumor-containing tests, we can propose a solution: We can run the algorithm with 2 different spatial decimation factor both in acceptable range. If algorithm points to different locations as ROIs, the test set does not contain tumor and artifacts have been located indeed but if the algorithm points to same locations, results are confirmed. The mentioned approach, plus applying the proposed framework on other microwave imaging methods will be presented in a future work of authors.

\section{Conclusion}
In this paper we proposed a novel computational complexity reduction framework on traditional beamforming techniques. We define two different modes for controlling the amount of complexity reduction. We applied our framework with proposed automatic and manual modes on DAS and DMAS techniques as the reference beamforming methods using two different datasets which both have tumors in different locations. The numerical analysis shows that our framework can reduce the complexity of DAS and DMAS methods and simultaneously detect the tumor location, correctly. It is also worth mentioning that the precision of tumor detection in our framework is completely dependent on the reference beamforming technique providing right selection of spatial decimation factor. So, it is guaranteed that in automatic mode, all tumors larger than the desired detectable tumor can be located as long as the reference beamforming is able to do so. This framework can be applied on more complex beamforming techniques, such as MVDR or MIST testing more complex datasets having skin and clutters. We aim at seeking for solutions to identify between clutter and tumor, as well as malignant tumor and benign tumor and also a much robust method to specify healthy patient from cancerous one in our future works.

\vspace{-5mm}
\small

\bibliographystyle{IEEEtran}
\bibliography{IEEEabrv,refs_Mohammad}

\end{document}